\documentclass[letterpaper, 10 pt, conference]{ieeeconf} 
\IEEEoverridecommandlockouts                              

\usepackage[T1]{fontenc}
\usepackage{amsmath,amsfonts}
\usepackage{array}
\usepackage{cite}
\usepackage[caption=false,font=scriptsize]{subfig}
\usepackage{url}
\usepackage{graphicx}
\usepackage[hidelinks]{hyperref}
\usepackage[none]{hyphenat}
\usepackage[capitalize]{cleveref}
\usepackage{tabularx}
\usepackage{booktabs}
\usepackage{siunitx}
\usepackage{blindtext}
\usepackage{glossaries}
\usepackage{subfig} 
\usepackage{multirow}
\usepackage[linesnumbered]{algorithm2e}
\RestyleAlgo{ruled}
\DontPrintSemicolon 

\usepackage{tikz}
\usetikzlibrary{arrows.meta}
\usetikzlibrary{arrows}
\usepackage{pgfplots}
\pgfplotsset{compat=1.16}
\usetikzlibrary{calc}
\usetikzlibrary{shapes,snakes}
\pgfdeclareplotmark{mystar}{
    \node[star,,star points=5, %
        star point ratio=0.5,
          inner sep=0pt,
          draw=black,
          solid, fill=black,
          minimum size=3pt
          ] {};
}
\usepackage{xcolor}
\definecolor{TUMBlue}{HTML}{0065BD}
\definecolor{DarkBlue}{HTML}{005293}

\definecolor{Orange}{HTML}{E37222}
\definecolor{Green}{HTML}{A2AD00}

\definecolor{Black}{HTML}{000000}
\definecolor{White}{HTML}{ffffff}

\definecolor{Gray}{HTML}{808080}
\definecolor{Graylight}{HTML}{CCCCCC}

\title{\LARGE \bf MP-RBFN: Learning-based Vehicle Motion Primitives\\using Radial Basis Function Networks}
\author{Marc Kaufeld$^{1}$, Mattia Piccinini$^{1}$ and Johannes Betz$^{1}$
\thanks{$^{1}$ All authors are with the Professorship of Autonomous Vehicle Systems, Technical University of Munich, 85748 Garching, Germany; Munich Institute of Robotics and Machine Intelligence (MIRMI). Contact: \{marc.kaufeld, mattia.piccinini, johannes.betz\}@tum.de}
}

\makeglossary

\begin{document}
\maketitle

\begin{abstract}
This research introduces MP-RBFN, a novel formulation leveraging \underline{R}adial \underline{B}asis \underline{F}unction \underline{N}etworks
for efficiently learning \underline{M}otion \underline{P}rimitives derived from optimal control problems for autonomous driving. 
While traditional motion planning approaches based on optimization are highly accurate, they are often computationally prohibitive.
In contrast, sampling-based methods demonstrate high performance but impose constraints on the geometric shape of trajectories. 
MP-RBFN combines the strengths of both by coupling the high-fidelity trajectory generation of sampling-based methods with an accurate description of vehicle dynamics.
Empirical results show compelling performance compared to previous methods, achieving a precise description of motion primitives at low inference times.
MP-RBFN yields a seven times higher accuracy in generating optimized motion primitives compared to existing semi-analytic approaches.
We demonstrate the practical applicability of MP-RBFN for motion planning by integrating the method into a sampling-based trajectory planner.
MP-RBFN is available as open-source software: \\\url{https://github.com/TUM-AVS/RBFN-Motion-Primitives}
\end{abstract}


\section{Introduction}
Real-time trajectory planning for autonomous vehicles (AVs) in dynamic environments, such as public roads shared with other traffic participants, is challenging due to the need to account for the motion of surrounding vehicles.
In such scenarios, optimal control (OC)-based planners are typically computationally prohibitive due to the non-convexity of collision avoidance constraints. 
On the other hand, graph- and sampling-based planners have recently gained popularity \cite{Trauth2024,Piazza_MPT_2024,Rowold2022}, relying on the construction and evaluation of a tree of motion primitives (MPs) to find the best trajectory. MPs are basic units of motion providing path and velocity information, that can be combined to form complex maneuvers. They have been developed for trajectory planning tasks in robotics \cite{Saveriano2023}, drones \cite{Mueller2015} and AVs \cite{Werling2010}. Moreover, MP-based planners can be integrated into modular differentiable perception-planning-control (PPC) stacks \cite{Karkus2022}, where the differentiability of the MPs is needed for end-to-end training of the entire system.
The design of MPs directly affects the planning performance and computational complexity. However, most of the existing MP formulations are limited by at least one of the following aspects: computational efficiency, dynamic feasibility, interpretability, or differentiability. 

This paper presents new motion primitives to learn the vehicle maneuvers computed by optimal control problems (OCPs). Our MPs are based on a generalized formulation of radial basis function networks (RBFNs), 
which are known to be universal function approximators and exhibit accurate interpolation properties \cite{rbf, rbf3}. 
Our results indicate that MP calculation based on RBFNs outperforms other MP calculation methods in terms of accuracy and differentiability, while maintaining competitive  inference times.

\begin{figure}[!t]
	\centering
    \includegraphics[width=0.97\columnwidth,trim=175 165 150 171, clip]{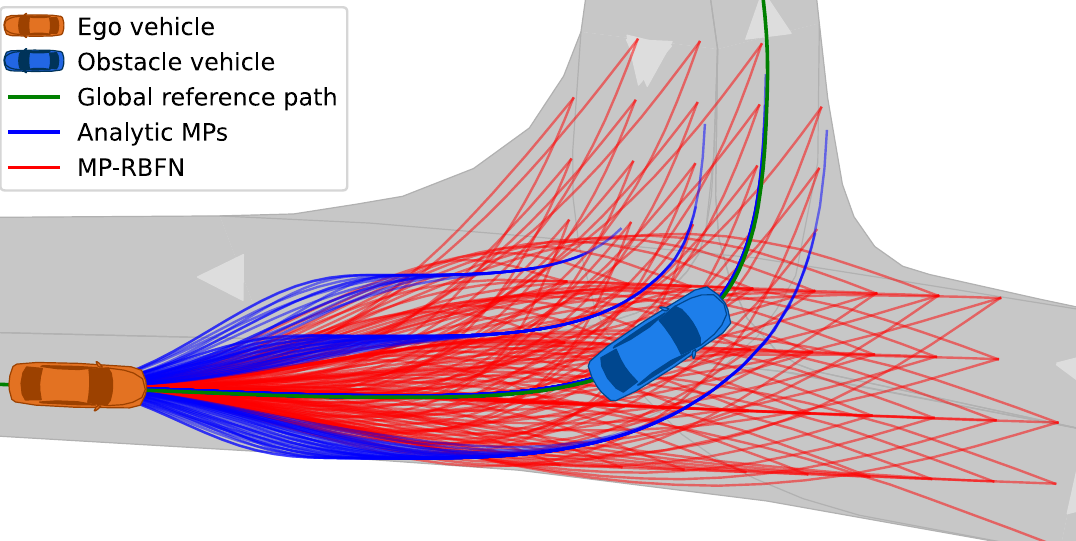}
    \caption{Comparison of analytic and MP-RBFN motion primitives in an intersection scenario.
The analytic MPs are jerk-minimal polynomials defined in a road-centered reference frame, which limits the exploration space as they naturally align with the reference path. In contrast, our MP-RBFN generates more diverse trajectories with lower computational cost, enabling a broader exploration of vehicle maneuver within the same re-planning time.
    }%
    \label{fig:topright}
\end{figure}%

\subsection{Related Work}
In the literature, methods to compute MPs for vehicle trajectory planning fall into three main groups: optimization-based, geometry-based, and learning-based techniques.
\begin{figure*}[!t]
	\centering
    \vspace*{2mm}
    \includegraphics[width=\textwidth]{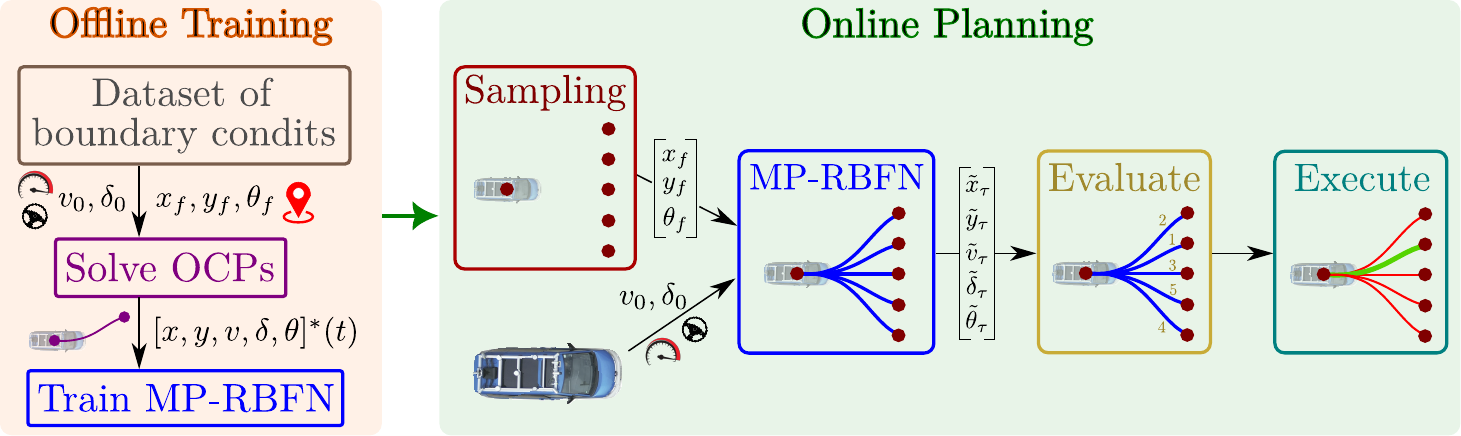}
	\caption{Overview of our framework. Offline, we train the MP-RBFN to learn the vehicle trajectories resulting from a custom OCP. Online, the trained MP-RBFN is used in a sampling-based motion planner, providing accurate and computationally efficient MPs. %
    }%
	\label{fig:framework}
\end{figure*}
\subsubsection{Optimization-based}
Optimal control problems can compute smooth and efficient vehicle maneuvers that minimize a given criterion, such as time \cite{Piccinini2024_how_optimal,Piccinini2024_ggv} or jerk~\cite{huang2023}, while complying with vehicle kinematic and dynamic constraints. However, the online evaluation of many OCP trajectories can be computationally prohibitive, especially in rapidly changing scenarios with non-convex dynamic obstacle avoidance constraints. Thus, some researchers devised simplified OC-based formulations for MPs 
aiming to approximate optimal solutions without the need to solve complex OCPs in real time.
For example, \cite{Lu2023}~employed a lookup table of MPs for online drift maneuver planning. 
Building on \cite{Werling2010} and \cite{DaLio2015}, analytic solutions of one-dimensional minimum-jerk OCPs have been applied to motion planning for AVs \cite{Rowold2022,Trauth2024} and drones \cite{Mueller2015,Richter2016}.
While these motion primitives offer computational efficiency and closed-form solutions, they are derived from simplified point-mass models with linearized vehicle dynamics and may lack dynamic feasibility in complex evasive maneuvers.

\subsubsection{Geometry-based}
Some authors used geometric curves such as cubic polynomials \cite{Stahl2019}, clothoids \cite{Frego2016,Brezak2014,Piazza_MPT_2024}, and Bézier curves \cite{Choi2010} for vehicle path planning. These curves are often used in graph- or sampling-based planners to connect pairs of waypoints since they can be efficiently calculated and do not require any training \cite{Stahl2019,Piazza_MPT_2024}. However, geometric curves do not explicitly  consider the vehicle dynamics and can therefore be locally inaccurate approximations of the OCP trajectories. 
Moreover, some geometric curves, like $G^2$ clothoids \cite{Bertolazzi2018G2}, may not be easily integrated into a differentiable PPC stack, as their computation requires numerical optimizations.

\subsubsection{Learning-based}
In general, learning-based methods can serve as black-box motion planners and have demonstrated good performance. However, they require lots of training data and often lack interpretability in their decision-making process \cite{bojarski2016, bansal2018}. 
Other authors proposed learning-based MPs as computationally efficient approximations of OCP solutions. Neural networks (NNs) were used in \cite{Gottschalk2024} to learn minimum-jerk OCP to control yaw accelerations, yet neglecting the longitudinal vehicle dynamics. In \cite{Piccinini2024_PathPoly}, polynomial-based NNs were proposed to learn the minimum-time OCP path, but the vehicle speed profile was not learned. 
Finally, RBFNs have been successfully used for approximating OCP solutions \cite{DeMarchi2022, Mirinejad2021} and for generating trajectories for manipulators \cite{Chettibi2018,Nadir2022}. In vehicle dynamics, to the best of our knowledge, RBFNs are only used for motion control \cite{Khai2019, Wang2022,Xiao2022}, but not for trajectory planning.

\subsubsection{Critical Summary}
To the best of our knowledge, the existing MP formulations are limited by at least one of the following aspects:
\begin{itemize}
    \item Numerical OCP-based MPs are accurate but computationally prohibitive in dynamic environments. 
    \item Analytical OCP-based MPs assume linear vehicle dynamics and may lack feasibility in high-speed or high-acceleration maneuvers.
    \item Geometric MPs are computationally efficient but neglect the vehicle's speed and dynamics.
    \item NNs were only used to learn the vehicle path resulting from OCPs, but not trajectories with detailed state vectors. 
    \item RBFNs were not used in the context of learning motion primitives used for sampling-based planning.
\end{itemize}

\subsection{Contributions}
To overcome the previously mentioned limitations of the existing literature, we propose the following contributions:
\begin{itemize}
    \item We present a new RBFN-based motion primitive formulation (\textit{MP-RBFN}) to efficiently learn the vehicle trajectories resulting from a custom user-defined OCP. 
    \item Our MP-RBFN is shown to achieve better accuracy compared to standard multilayer perceptrons (MLPs) and a basic RBFN at compelling training times. 
    \item We discuss how our MP-RBFN yields lower inference times with respect to the original OCP and to minimum-jerk semi-analytical primitives \cite{Werling2010}.
    \item We present an application example of MP-RBFN in a sampling-based planner to compute dynamic obstacle avoidance maneuvers. Also, we discuss a possible integration of MP-RBFN into a fully-differentiable PPC software stack.
\end{itemize}

\section{Methodology}

\begin{figure}[!ht]
	\centering
    \vspace*{2mm}
    \includegraphics[width=0.96\columnwidth]{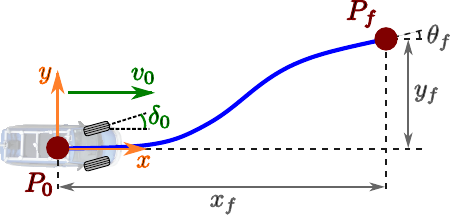}
	\caption{The proposed MP-RBFN learns the OCP trajectory to connect an initial waypoint $P_0$ with a final waypoint $P_f$. The MP-RBFN's inputs are the 
    initial velocity $v_0$ and steering angle $\delta_0$, and the coordinates $\{x_f,y_f\}$ and  yaw angle $\theta_f$ at the final position. 
    }%
	\label{fig:RBF_input_output}
\end{figure}

Fig. \ref{fig:framework} provides an overview of our framework.
Offline, we train our MP-RBFN to learn the vehicle trajectories resulting from a custom OCP. Specifically, MP-RBFN learns the optimal maneuvers connecting an initial waypoint $P_0$ to a final waypoint $P_f$ over a fixed time horizon $T$ (Fig. \ref{fig:RBF_input_output}).

We define a data set of boundary conditions for which we solve the OCP numerically to find kinematically and dynamically feasible trajectories. We use these trajectories to train our MP-RBFN. 
The trained MP-RBFN can be integrated into a sampling-based motion planner in the online operation phase. In each planning step, the  planner computes and evaluates a set of MPs, and selects the best one to follow.  
In the following, we first describe the OCP used for training and then outline the architecture of MP-RBFN.

\subsection{Optimal Control Problem }
\label{ssec:ocp}
\begin{figure}[!t]
\vspace*{1mm}
	\centering
    \includegraphics[width=0.65\columnwidth]{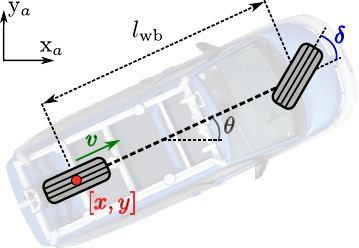}
	\caption{Single-track model with reference point on the rear axle \cite{Althoff.2020}.
    }%
	\label{fig:vehicle_model}
\end{figure}

First, we want to emphasize that our approach does not rely on a specific model formulation and may be applied to any OCP variant for trajectory optimization. In our example, the OCP returns jerk-minimal trajectories.
To model the vehicle dynamics, we employ a non-linear kinematic single-track model \cite{Althoff.2020}, as shown in Fig. \ref{fig:vehicle_model}. The state vector $\mathbf{x} = [x,y, \delta, v, a, \theta]$ encompasses the $x$- and $y$-coordinates of the rear axle in the absolute reference frame $(x_a ,\, y_a)$, the steering angle $\delta$, velocity $v$, acceleration $a$ and the yaw angle $\theta$.
The control inputs $\mathbf{u} = [j_{\mathrm{long}}, v_\delta]$ comprise the longitudinal jerk $j_{\mathrm{long}}$ and the steering angle velocity $v_\delta$. The time-domain model equations are:
\begin{equation}
    \label{eq:dyn}
    \dot{\mathbf{x}}(t) = 
    \begin{bmatrix}
        \dot{x}(t)\\
        \dot{y}(t) \\
        \dot{\delta}(t) \\
        \dot{v}(t)\\
        \dot{\theta}(t)    \\
        \dot{a}(t) \\
    \end{bmatrix} = 
    \begin{bmatrix}
        v(t) \, \cos{\theta(t)}\\
        v(t) \, \sin{\theta(t)}\\
        v_\delta(t)\\
        a(t)\\
       \frac{v(t)}{l_{\mathrm{wb}}} \tan \delta(t) \\
        j_{\mathrm{long}}(t) \\
    \end{bmatrix}= \mathbf{f}\big(\mathbf{x}(t),\mathbf{u}(t)\big),
\end{equation}
where $l_{\mathrm{wb}}$ is the vehicle wheelbase.
In addition, we constrain the feasible vehicle states and controls:
\begin{equation}
    \label{eq:cx}
    \mathbf{c}\big(\mathbf{x}(t),\mathbf{u}(t)\big) = 
    \begin{bmatrix}
        |\delta(t)| -\delta_{\mathrm{max}} \\
        v(t) - v_{\mathrm{max}} \\
        -v(t) -v_{\mathrm{min}} \\
        |v_\delta(t)| - v_{\delta_{\mathrm{max}}} \\
        \left(\frac{a(t)}{\bar{a}_{\mathrm{long}}(a,v)}\right)^2 + \left(\frac{\dot{\theta}(t) v(t)}{a_{\mathrm{max,lat}}}\right)^2  -1\\
    \end{bmatrix}
    \leq \mathbf{0}.
\end{equation}
The steering angle $\delta$ is limited by the mechanical bound $\delta_{\mathrm{max}}$, while the velocity is constrained between $0$ and $v_{\mathrm{max}}$. The steering rate is capped at $v_{\delta{\mathrm{max}}}$, and the longitudinal-lateral accelerations must lie within an elliptic g-g diagram. Engine power limitations are approximated by heuristically reducing the longitudinal acceleration bound $\bar{a}_{\mathrm{long}}(v)$ at higher velocities:

\begin{equation}
    \label{eq:acc}
   \bar{a}_{\mathrm{long}}(a,v) = \begin{cases}
       a_{\mathrm{max, long}} \frac{v_s}{v(t)} & \text{if} \; v(t) > v_s\\
       -a_{\mathrm{max, long}} & \text{if} \; a(t) < 0 \\
       a_{\mathrm{max, long}} & \text{otherwise},
   \end{cases} 
\end{equation}
where $v_s$ is the velocity after which the achievable acceleration starts decreasing due to driving resistances \cite{Althoff.2020}.

Given the vehicle dynamics $\mathbf{f}(\cdot)$ in \eqref{eq:dyn} and the set of constraints $\mathbf{c}(\cdot)$ in \eqref{eq:cx}, we define the following OCP to find the jerk-minimal trajectory $\mathbf{\zeta}=\{\mathbf{x}^*(t), \mathbf{u}^*(t), t \in T\}$ over a fixed time horizon $T$:

\begin{align}
   \min\; &\mathcal{L} = \int_0^T w_1 j_{\mathrm{long}}(t)^2 + w_2 j_{\mathrm{lat}}(t)^2\, \mathrm{d}t \label{eq:ob} \\
     \text{s.t.}&  \nonumber \\
    &\dot{\mathbf{x}}(t) = \mathbf{f\big(x(t),u(t)\big)}  \label{eq:sp}\\ 
     & \mathbf{c}\big(\mathbf{x}(t),\mathbf{u}(t)\big) \leq \mathbf{0} \label{eq:cxu} \\     
    &\boldsymbol{\psi}\big(\mathbf{x}(0) - \mathbf{x}_0\big) = \mathbf{0} \label{eg:ini}\\
    &\boldsymbol{\psi}\big(\mathbf{x}(T)\big) = \mathbf{0} \label{eq:fin}\\
    &\mathbf{u}(T) = \mathbf{0}.  \label{eq:u}
\end{align}
The cost function $\mathcal{L}$ in \eqref{eq:ob} is the weighted integral over the time of the squared lateral and longitudinal jerk. The jerk in vehicle direction is directly given as a control input while the lateral jerk can be derived from the lateral acceleration:
\begin{equation}
    j_{\mathrm{lat}} = 
    \frac{d}{dt}\left(\dot{\theta} v\right) = \frac{2v\dot v}{l_{\mathrm{wb}}}\tan\delta + \frac{v^2}{l_{\mathrm{wb}}\cos^2 \delta }\dot\delta.
\end{equation}

We select the weights $w_{1,2}$ such that the controls are normalized with respect to their maximal allowed values.
The initial and final states $\{\mathbf{x}_0, \mathbf{x}_f\}$ are imposed by \eqref{eg:ini} and \eqref{eq:fin}, in the following way:
\begin{equation}
    \label{eq:bcs}
    \begin{bmatrix}
        x(0)\\
        y(0)\\
        \delta(0)-\delta_0\\
        v(0) - v_0\\
        \theta(0)    
    \end{bmatrix} = \mathbf{0} \quad \text{and} \quad
    \begin{bmatrix}
        x(T) - x_f\\
        y(T) - y_f\\
        \delta(T) \\
        a(T) \\
        \theta(T) - \theta_f
    \end{bmatrix} = \mathbf{0}.
\end{equation}
Without loss of generality, we solve the OCP in a local coordinate system with the origin at the vehicle's rear axle and oriented along the vehicle center line. 
Note that the initial acceleration $a_0$ and the velocity at the final waypoint $v_f$
are not constrained but output of the optimization.
This simplifies the initial condition to a zero vector except for the initial velocity and steering angle.
The calculated MPs can then be used for local motion planning in vehicle-centered coordinates, simplifying the maneuver calculations.
Finally,~\eqref{eq:u} sets the final control values to zero. 
\subsection{Proposed Network Architecture}
This section outlines the design of the proposed MP-RBFN, starting from its inputs and outputs and then moving on to the internal architecture.
\subsubsection{Inputs and Outputs} \label{sec:input_output}
As shown in Fig. \ref{fig:RBF_input_output}, our MP-RBFN learns the numerical OCP trajectories in the time domain from an initial waypoint $P_0$ to a final one $P_f$, within a fixed planning horizon $T$. The MP-RBFN's input vector~$\mathbf{q}  \in \mathbb{R}^{5}$ comprises the velocity $v_0$ and steering angle~$\delta_0$ in $P_0$, and the position $\{x_f, y_f\}$ and yaw angle~$\theta_f$ in $P_f$: 

\begin{align}
     \mathbf{q} & =
    \begin{bmatrix}
       v(0) &  \delta(0) & x(T) & y(T) & \theta(T)
    \end{bmatrix}^\mathsf{T} \label{eq:input_q} \\
        & =
    \begin{bmatrix}
       v_0 &  \delta_0 & x_f & y_f & \theta_f
    \end{bmatrix}^\mathsf{T}, \nonumber
\end{align}
where all the quantities are expressed in the vehicle-centered reference frame.
The output of our RBFN is a discretized trajectory 
$\tilde{\boldsymbol{\zeta}}=\{\tilde{\mathbf{x}}_\tau,  \tau \in  \boldsymbol{t}_o\}$,
with coordinates~$\{\tilde{x}_\tau,\tilde{y}_\tau\}$, velocities $\tilde{v}_\tau$, steering angles~$\tilde{\delta}_\tau$ and yaw angles~$\tilde{\theta}_\tau$, discretized at the time steps $\tau \in \boldsymbol{t}_o$:

\begin{equation}
    \boldsymbol{t}_o = 
    \begin{bmatrix}
        0 & \tau_1 & \tau_2 & \dotsc & T
    \end{bmatrix}^\mathsf{T}.
    \label{eq:time_vector}
\end{equation}

\subsubsection{Internal Architecture}
%
Our MP-RBFN employs a new extended RBFN formulation, with an additional input layer to improve the model's accuracy and accelerate the training convergence. 
First, we will revise the basic RBFN architecture, followed by an introduction to our proposed MP-RBFN.
%
\paragraph{Basic RBFN}
RBFNs are feedforward neural networks that use  radial basis functions (RBFs) as activation functions for hidden neurons. 
They have a three-layer architecture: an input layer receives the input vector $\mathbf{q} \in \mathbb{R}^n $, a hidden layer with $K$ neurons and RBFs as activation functions, and a linear output layer, mapping the input vector to a scalar value. Mathematically, RBFNs define a mapping $\varphi:\mathbb{R}^n \rightarrow \mathbb{R}$ such that:
\begin{equation}
    y = \varphi(\mathbf{q}) = \sum_{k=1}^K w_k \, \rho\big(\epsilon||\mathbf{q}-\mathbf{c}_k ||\big).
\end{equation}
The RBF activation functions are radially symmetric: their output is based on the distance $r =||\mathbf{q}-\mathbf{c}_k ||$ between the input $\mathbf{q} \in \mathbb{R}^n$ and the center point $\mathbf{c}_k$ of neuron $k\in K$.
The function $\rho(\cdot)$ is the RBF kernel which determines the shape of the activation function and $\epsilon$ is a shape parameter. Common choices for $\rho(\cdot)$ are Gaussian or inverse (multi-)quadratic functions.  
We tested different activation functions and state the results in the \cref{ssec:train}. 
Note that the region of influence for each center $\mathbf{c}_k$ is limited, since $\lim\limits_{r \to \infty} \rho(r) = 0$. Finally, the trainable weights $\boldsymbol{w} \in \mathbb{R}^K$ control the impact of each RBF neuron on the output.

In \cite{lowe.1988} it has been demonstrated that RBFNs can learn multi-dimensional variables by expanding the dimensionality of the weight matrix $\boldsymbol{W} \in \mathbb{R}^{I \times K}$:
\begin{equation}
    y_i=\varphi_i(\mathbf{q}) = \sum_{k=1}^K W_{ik} \, \rho\big(\epsilon||\mathbf{q}-\mathbf{c}_k ||\big) \quad \forall i \in I.
    \label{eq:rbfn_base}
\end{equation}
\paragraph{Proposed MP-RBFN}
\begin{figure}[!ht]
	\centering
    \includegraphics[width=1\columnwidth]{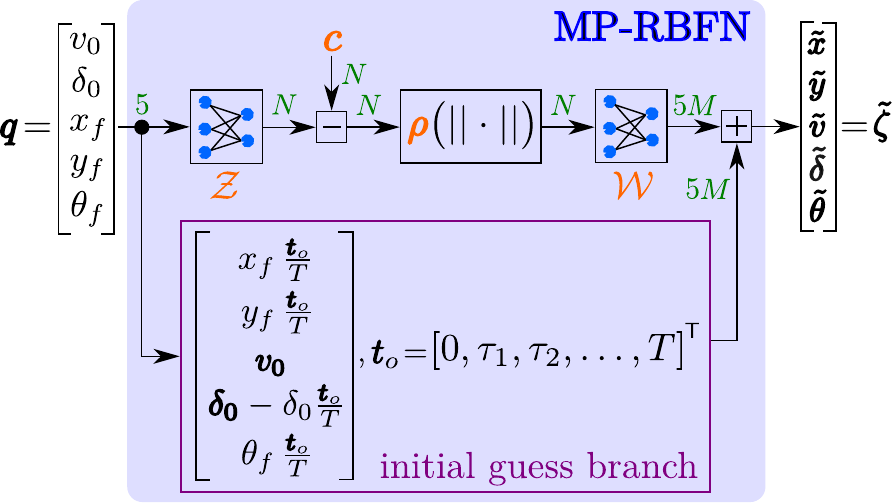}
    \caption{Architecture of the proposed MP-RBFN. The learnable parts are colored in orange, while the numbers in green indicate the dimensions of the corresponding quantities. 
    }%
	\label{fig:MP_RBFN_architecture}
\end{figure}%
The architecture of the proposed MP-RBFN is depicted in Fig. \ref{fig:MP_RBFN_architecture} and is given by:

\begin{align}
    \label{eq:our_MP_RBFN}
     \!\!
    \tilde{\boldsymbol{\zeta}}\!\!
    = \!\!\begin{bmatrix}
        \tilde{x}_\tau  \\
        \tilde{y}_\tau \\
        \tilde{v}_\tau \\
        \tilde{\delta}_\tau \\
        \tilde{\theta}_\tau
    \end{bmatrix}_{\tau \in \boldsymbol{t}_o}
    \!\!\!\!\!\!\!\!\!\!\!\! = \sum_{k=1}^K W_{\tau,k} \rho\Big(\epsilon \big\|\underbrace{\boldsymbol{Z} \mathbf{q}}_{z \in \mathcal{Z}} - \mathbf{c}_k \big\|\Big)
    \!+ \!\underbrace{
    \begin{bmatrix}
        x_f  \frac{\boldsymbol{t}_o}{T} \\
        y_f  \frac{\boldsymbol{t}_o}{T} \\
        \boldsymbol{v_0} \\
        \boldsymbol{\delta_0}-\delta_0 \frac{\boldsymbol{t}_o}{T} \\
        \theta_f  \frac{\boldsymbol{t}_o}{T}
    \end{bmatrix}}_{\text{init. guess}}\!\!.
\end{align}

In comparison to \eqref{eq:rbfn_base}, we use an additional linear layer with weights $\boldsymbol{Z} \in \mathbb{R}^{N \times n}$ to map the input vector $\mathbf{q} \in \mathbb{R}^{n}$ to the latent space $\mathcal{Z} \in \mathbb{R}^N$ (Fig. \ref{fig:MP_RBFN_architecture}). In our case, the size of the input $\mathbf{q}$ is $n=5$ (\cref{eq:input_q}), while $N$ is a hyperparameter that determines the dimension of the latent space.
Each latent variable in $z \in \mathcal{Z}$ is then processed by the RBF layer. The network outputs a trajectory $\tilde{\boldsymbol{\zeta}}\in \mathbb{R}^{5 \times M}$ with $M$ the number of discretized timesteps $\tau$.

Moreover, to satisfy the initial and final conditions from~\cref{eq:bcs} and provide a meaningful initialization for the training, we test an additional additive branch in \cref{eq:our_MP_RBFN}. This branch performs a linear interpolation in the time domain between the known initial and final states, using the time vector $\boldsymbol{t}_o$. 
For the $x$- and $y$-coordinates as well as the heading, the initial values are set to zero, simplifying the interpolation as stated.
Since we do not impose a final condition on the velocity $v$, the initial guess is a constant vector $\boldsymbol{v_0}$, repeating the initial velocity across all time steps.
For the steering angle $\delta$, we assume a non-zero initial value and a final value of zero at the end of the planning horizon $T$. 
$\boldsymbol{\delta_0}$ is therefore a vector of the initial steering angle repeated over all time steps, and the second term $-\delta_0 \frac{\boldsymbol{t}_o}{T}$ linearly reduces $\delta$ to zero by the final time step.

\subsubsection{Learnable Parameters}
There exist three possibilities to train RBFNs \cite{rbf2}: 
In one-step learning, only the output weights $\boldsymbol{W}$ are trained to match a desired output, while the center points $\mathbf{c}$ and shape parameters $\epsilon$ remain fixed. 
In two-step learning, first  $\mathbf{c}$ and $\epsilon$ are optimized by unsupervised learning methods, followed by the optimization of the weights $\boldsymbol{W}$.
Finally, RBFNs can be tuned through backpropagation, similarly to the training of MLPs, where $\mathbf{c}$, $\boldsymbol{W}$ and $\epsilon$ are trainable parameters. We adopt this last method to train our MP-RBFN.
\section{Experiments \& Results}
\label{sec:exp}
In this section, we first describe how the dataset of OCP motion primitives was created, followed by the training procedure and final results. We compare the accuracy and inference time with benchmark NNs and semi-analytic MPs.

\subsection{Dataset Creation}
To create the dataset of MPs for the MP-RBFN, we solve the OCP introduced in  \cref{ssec:ocp} for a multitude of different initial and final conditions. The vehicle parameters are based on a medium-sized vehicle of type BMW 320i \cite{Althoff.2020} and are listed in \cref{tab:veh}. To account for inaccuracies of the simplified bicycle model at higher lateral accelerations, we adopt the recommendation from \cite{Polack.2017} and limit the lateral acceleration to \SI{0.5}{g}.

\begin{table}[!htb]
\centering
\caption{Parameters of the vehicle dynamics model \cite{Althoff.2020}.}
\renewcommand{\arraystretch}{1.15}
\begin{tabularx}{0.75\linewidth}{l l l}
    \toprule
    \textbf{Parameter} & \textbf{Notation} & \textbf{Value} \\
    \midrule
    Wheelbase & $l_{\mathrm{wb}}$& \SI{2.6}{\meter}\\
    Max steering angle & $\delta_{\mathrm{max}}$ & \SI{1.0}{\radian}\\
    Max steering rate & $v_{\delta_{\mathrm{max}}}$ & \SI{0.4}{\radian\per\second}\\
    Max velocity & $v_{\mathrm{max}}$ & \SI{28}{\meter\per\second}\\
    Max long. acceleration & $a_\mathrm{max, long}$ & \SI{11.5}{\meter\per\square\second} \\
    Max lat. acceleration & $a_\mathrm{max, lat}$ & \SI{4.9}{\meter\per\square\second} \\
    Switching velocity & $v_s$  & \SI{7.4}{\meter\per\second} \\
    \bottomrule
\end{tabularx}
\label{tab:veh}
\end{table}

The initial position and yaw angle are fixed to be zero, $\{x_0,y_0\} = \{\SI{0}{\meter},\SI{0}{\meter}\}$ and $\theta_0=\SI{0}{\radian}$, respectively.
The initial velocity $v_0$ is varied over the range $v_0 \in [\SI{0}{\meter\per\second}, \SI{28}{\meter\per\second}]$ with a step size of \SI{1}{\meter\per\second}. 
Since the motion primitives are to be used for reactive short term motion planning in a receding horizon technique, the planning horizon is fixed to a time interval $T=\SI{3}{\second}$ with a discretization of $\Delta t = \SI{0.1}{\second}$.
The final positions are all points $\{x_f,y_f\}$ within the range $[r_\mathrm{min},r_\mathrm{max}]$ covered by the maximal de- and acceleration $\bar{a}_{\mathrm{long}}(a,v)$  for a given initial velocity $v_0$.

\begin{align}
r_{\mathrm{min}}(v_0) &= \frac{v_0^2}{2a_{\mathrm{min}}} \\
r_{\mathrm{max}}(v_0) &= \frac{1}{2} \bar{a}_{\mathrm{long}}(a,v)T^2 + v_0 T.
\end{align}
The step sizes are \SI{3}{\meter} and \SI{1}{\meter} in the $x$- and $y$ directions, respectively.
To also cover motion primitives for turning maneuvers, the final yaw angle is altered in \SI{0.16}{\radian} steps between $\delta_{f,\mathrm{min}} = \SI{-1.6}{\radian}$ to $\delta_{f,\mathrm{max}} = \SI{1.6}{\radian}$. This range covers trajectories driving around \SI{90}{\degree} curves in both directions.
The initial steering angle is altered within the feasible range $|\delta_{\mathrm{max}}|<\SI{1}{\radian}$ with a discretization of $\Delta \delta = 0.1$.
Since not all combinations are kinematically feasible or reachable within the specified time horizon, infeasible combinations without a valid OCP solution are neglected. 
We use the open-source optimization tool CasADI \cite{Andersson2019} to solve the OCP.
The data set consists of a total of approximately $2.6$ million MPs used to train MP-RBFN. 

\subsection{Training}
\label{ssec:train}
The model is implemented in PyTorch. We set the size of the latent space $N$ and the number of hidden neurons with trainable centers $K$ to $1024$. 
The final layer outputs a vector of size $I=155$, discretizing the five states of the trajectory  $\tilde{\boldsymbol{\zeta}}$  with $\Delta\tau=\SI{0.1}{\second}$ over a planning horizon of $T=\SI{3}{\second}$.
With the initial and final linear layers, this results in a total of $165k$ trainable parameters. 
To balance different scales in the output values during the training, we use the weighted sum of the mean square errors (MSE) of the position, velocity and orientation between the predicted trajectory and the OCP solution as loss function. The Adam optimizer~\cite{adam} is used for optimization with a learning rate of $1\cdot 10^{-3}$. As hyperparameters for the optimizer, we adhere to the suggested values $\beta_1=0.9$ and $\beta_2=0.999$.
MP-RBFN is trained for $2.000$ epochs with a training and test data split of \SI{70}{\percent} and \SI{30}{\percent}, respectively. 

We test three different RBFs as activation functions for the hidden layer. The minimal losses on the test split are listed in \cref{tab:actfunc}. While inverse quadratic and inverse multiquadratic RBFs perform comparable with MSEs of $0.36$ and $0.34$, respectively, the Gaussian activation function yields better results with a loss of $0.24$.
In the subsequent evaluations, we therefore only refer to the MP-RBFN with Gaussian activation function.
\begin{table}[!hb]
\centering
\caption{Test set performance across RBF activation functions.}
\renewcommand{\arraystretch}{1.15}
\begin{tabularx}{0.85\linewidth}{l  c c c}
    \toprule
   \multirow{2}{*}{\textbf{}} &  \textbf{Gaussian} & \textbf{Inv. Quadratic}   &  \textbf{Inv. Multiquadratic}    \\
   & $\mathrm{e}^{-(\epsilon r)^2}$ & $\frac{1}{1+(\epsilon r)^2 }$ &  $\frac{1}{\sqrt{1+(\epsilon r)^2} }$\\
    \midrule
    MSE & $0.24$ & $0.36$ & $0.34$ \\
    \bottomrule
\end{tabularx}
\label{tab:actfunc}
\end{table}

\subsection{Benchmarking}
In this section, we evaluate the performance of MP-RBFN. 
As a benchmark, we train 4 additional models with the same settings and the same dataset: 
MP-RBFN without interpolating branch (MP-RBFN\textsubscript{w/o interp}), two MLPs and a standard RBFN as defined in \cref{eq:rbfn_base}.
The MLPs consist of three fully connected  linear layers  with $1024$ hidden neurons, resulting in a total of $164k$ parameters. One uses tanh (MLP\textsubscript{tanh}) and the other sigmoid functions (MLP\textsubscript{sig}) as activation function.
To keep the number of trainable parameters comparable, the basic RBFN also consists of $1024$ centers with Gaussian activation function, but since the first linear layer is omitted, the centers are vectors with the same size as the input, yielding a total of $165k$ parameters.
Additionally, analytical jerk-optimal MP calculation based on quintic polynomials in longitudinal and lateral direction  \cite{Trauth2024} is used for comparison.

First, we compare the accuracy of the predicted trajectories with the numerically optimized solution of the OCP for different final conditions. Second, we compare the computational efficiency by measuring the inference time to generate a tree of motion primitives.

The interpolating branch does not significantly improve the final accuracy of the model, as after the training the minimal loss on the test split is $0.24$ compared to $0.26$ for MP-RBFN\textsubscript{w/o interp} without interpolation. 
The average MSE on the test dataset for the MLP\textsubscript{sig} is $0.92$, while 
MLP\textsubscript{tanh} achieves a loss of $0.85$.
The basic RBFN  does not converge within the fixed number of epochs with a MSE of $21.5$. This model is therefore discarded in the subsequent benchmarking comparisons.

\begin{table}[!hb]
\centering
\caption{RMSE between OCP and predicted MP.}
\renewcommand{\arraystretch}{1.15}
\begin{tabularx}{0.95\linewidth}{l c c c}
    \toprule
   \textbf{Model} & 
   \begin{tabular}{@{}c@{}}\textbf{Position} \\ in \si{\meter}\end{tabular}
   & \begin{tabular}{@{}c@{}} \textbf{Velocity} \\ in \si{\meter\per\second} \end{tabular}
   & \begin{tabular}{@{}c@{}} \textbf{Orientation} \\ in \si{\radian} \end{tabular}\\
    \midrule
    Analytic MP 
    & $3.17$
    & $0.98$
    & $0.29$
    \\
    MLP\textsubscript{sig}
    & $0.51$ 
    & $0.31$
    & $\textbf{0.01}$
    \\
    MLP\textsubscript{tanh}
    & $0.49$
    & $0.34$
    & $ 0.03$
    \\
     \textbf{MP-RBFN\textsubscript{w/o interp} (ours)}
    & $ 0.24$
    & $\textbf{0.17}$
    & $0.02$
    \\
    \textbf{MP-RBFN (ours)}
     & $\textbf{0.23}$  
     & $\textbf{0.17}$ 
     & $0.02$\\
    \bottomrule 
\end{tabularx}
\label{tab:train}
\end{table}
The average root mean square error (RMSE) for the position, velocity profile and orientation along the MPs are shown in \cref{tab:train} for the test dataset. 
Also for the individual states, it is apparent that the additional interpolation branch of MP-RBFN does not improve the precision compared to MP-RBFN\textsubscript{w/o interp}.
Both MP-RBFN variants performs better than the MLPs and the analytic calculation.
On average, the predicted positions for MP-RBFN along the trajectory differ by \SI{0.23}{\meter}, which, with an average length of the trajectories of \SI{83}{\meter} corresponds to a deviation of \SI{0.3}{\percent}. The velocity and orientation yield an accuracy of \SI{0.17}{\meter\per\second} and \SI{0.02}{\radian}. 
The RMSEs for the position and velocity of the MLPs are twice as high as for MP-RBFN. While MLP\textsubscript{sig} achieves a smaller orientation error, MLP\textsubscript{tanh} performs slightly worse compared to our MP-RBFN.
The analytic benchmark method, although following a commonly used method for calculating analytical motion primitives, performs worst and finds kinematically feasible trajectories only for \SI{47}{\percent} of the final conditions in the training dataset. On average, these MPs differ most in position, velocity and orientation from the OCP solution.

\begin{table}[!htb]
\centering
\caption{Mean position error in \si{\meter} for different final yaw angles.}
\renewcommand{\arraystretch}{1.15}
\begin{tabularx}{0.95\linewidth}{l  c c c}
    \toprule
    \multirow{2}{*}{\textbf{Model}} 
    & \multicolumn{3}{c}{\textbf{Final yaw angle} }\\
     &  \SI{0}{\radian}
     & \SI{0.8}{\radian} 
     & \SI{1.6}{\radian}\\
    \midrule
    Analytic MP 
     &  \SI{1.40}{\meter}
     & \SI{4.06}{\meter} 
     & \SI{6.02}{\meter} 
      \\
     (valid share)
     & (\SI{60}{\percent})
     & (\SI{51}{\percent})
     & (\SI{12}{\percent})
     \\
     MLP\textsubscript{sig} 
     & \SI{0.47}{\meter}  
     & \SI{0.49}{\meter} 
     & \SI{0.63}{\meter} 
    \\
    MLP\textsubscript{tanh}
    & \SI{0.46}{\meter} 
    & \SI{0.47}{\meter} 
    & \SI{0.58}{\meter}
    \\    
    \textbf{MP-RBFN\textsubscript{w/o interp} (ours)}
    & \SI{0.23}{\meter} 
    & \SI{0.23}{\meter}
    & \SI{0.31}{\meter}
    \\
     \textbf{MP-RBFN (ours)}
     & $\textbf{0.22\,m}$ 
     & $\textbf{0.22\,m}$ 
     & $\textbf{0.30\,m}$  \\
    \bottomrule
\end{tabularx}
\label{tab:acc}
\end{table}

 In \cref{tab:acc}, the positional deviations for different final yaw angles are listed. 
 The proposed MP-RBFN, along with all benchmark networks, remains robust to variations in the final heading with only a degradation of approximately \SI{10}{\centi\meter} in the highest final yaw angles.
In contrast the accuracy of the analytic MPs deteriorates with increasing angle due to the limited geometric flexibility of the underlying polynomial formulation. At a final yaw angle of \SI{1.6}{\radian}, the mean position error along the trajectories  reaches \SI{6.02}{\meter}, which is over four times higher than the error at \SI{0}{\radian}. Interestingly, also for final states aligned with the initial orientation, the analytic approach performs seven times worse than our MP-RBFN.

For the comparison of the computational efficiency, we sample $1000$ times a set of trajectories and measure the inference time. The mean values are compared  in \cref{tab:inf_time}.
The evaluations are done on an NVIDIA GeForce RTX 5090 GPU and AMD Ryzen 9 7950X 16-core processors. For the analytic approach, we utilize parallelized multiprocessing to reduce the inference times on the CPU.
All approaches are implemented in Python.

\begin{table}[!hbt]
\centering
\caption{Inference times for sampling the stated number of trajectories. $(\cdot)^*$ indicates parallelized computation. Network times are on GPU (CPU in parentheses).}
\renewcommand{\arraystretch}{1.15}
\begin{tabularx}{0.99\linewidth}{l c c c}
    \toprule
    \multirow{2}{*}{\textbf{Model}} & \multicolumn{3}{c}{\textbf{Number of trajectories} }\\
    & \textbf{50 } 
    & \textbf{3500} 
    & \textbf{13000} \\
    \midrule
    OCP Solver
    & \SI{2.4}{\second} 
    & \SI{174.6}{\second} 
    & \SI{658.6}{\second} 
    \\
    Analytic MP
    & \num{35.7}~\si{\milli\second} 
    & $\num{0.6}~\si{\second}^*$ 
    & $\num{1.8}~\si{\second} ^*$ \\
    
    MLP\textsubscript{tanh/sig}
    & $\textbf{0.02}$~\si{\milli\second} 
    &  $\textbf{0.05}$~\si{\milli\second}  
    & $\textbf{0.25}$~\si{\milli\second}  
    \\
    & (\num{0.05}~\si{\milli\second})
    &(\num{1.7}~\si{\milli\second})
    &(\num{13.5}~\si{\milli\second})  
    \\
    \textbf{MP-RBFN\textsubscript{w/o interp} (ours)}
    & \num{0.04}~\si{\milli\second}
    & \num{0.09}~\si{\milli\second} 
    & \num{0.39}~\si{\milli\second}
    \\
    & (\num{0.06}~\si{\milli\second})
    & (\num{4.9}~\si{\milli\second})
    &(\num{30.1}~\si{\milli\second})
    \\
    \textbf{MP-RBFN (ours)}
    & \num{0.13}~\si{\milli\second} 
    & \num{0.21}~\si{\milli\second} 
    & \num{0.55}~\si{\milli\second}  
    \\
    & (\num{0.10}~\si{\milli\second})
    &  (\num{5.1}~\si{\milli\second})
    & (\num{31.4}~\si{\milli\second})
    \\
    \bottomrule
\end{tabularx}
\label{tab:inf_time}
\end{table}
\begin{figure*}[!t]
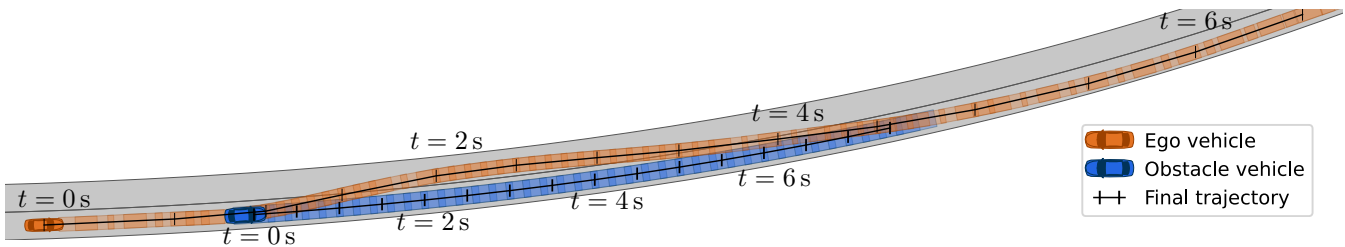

    \centering
        \include{figures/overtake}
    \caption{Overtaking maneuver with a sampling planner utilizing MP-RBFN motion primitives. The ego vehicle starts with $v_0=\SI{20}{\meter\per\second}$, while the obstacle vehicle drives at a constant velocity $v=\SI{10}{\meter\per\second}$. The markers on the trajectory correspond to a discretization in time of \SI{0.5}{\second}.
    }
    \label{fig:overtake}
\end{figure*}
The OCP solver performs worse followed by the analytic MP calculation which accelerates the inference time by a factor up to $10^2$. With a low number of trajectories, the multiprocessing implementation does not result in any improvements for the analytic approach. It is evident, that the inference times for all NN-based approaches are much lower, both on CPU and GPU, especially with increasing number of trajectories, since they are simple feedforward networks and do not rely on any optimization.
No difference is recognizable between the two MLP versions, while MP-RBFN\textsubscript{w/o interp} only needs marginally more computation time. MP-RBFN has the slowest inference among all trained models.  The added complexity in the network architecture through the interpolation results in higher inference times. 

\subsection{Application Example in a Sampling-based Motion Planner}

Finally, to illustrate its practical usability, we present an exemplary application of MP-RBFN in a sampling-based motion planner for autonomous driving.
The planner is implemented within the CommonRoad simulation environment \cite{althoff2017} and samples, evaluates and follows a MP-RBFN trajectory in a receding horizon technique with a step size of $\Delta t=\SI{0.1}{\second}$.
Given a global trajectory consisting of a reference path and a velocity profile calculated a priori through convex optimization \cite{shimizu2022}, the sampled trajectories are evaluated using a set of cost functions that consider the deviation from the desired velocity and the offset to the global trajectory.
Additionally, to take into account other traffic participants, collision costs based on predictions calculated with a neural network \cite{Geisslinger2021} are included. 
The trajectory with the lowest total costs is executed for one planning step and then the trajectories are resampled.
In \cref{fig:overtake}, we show an exemplary overtake maneuver performed with our planner. The blue vehicle drives with a constant velocity of \SI{10}{\meter\per\second} while the AV approaches with initially \SI{20}{\meter\per\second}. 
It can be seen, that by reevaluating MP-RBFN trajectories, the motion planner is able to perform a safe overtake maneuver with sufficient distance to the other vehicle.

\section{Discussion}
Our results demonstrate the effectiveness of the proposed MP-RBFN in generating motion primitives for vehicles from custom user-defined OCPs.
We varied our model with different activation functions and tested the value of an additional interpolation branch. While changing the activation function to a Gaussian improved the accuracy, the added interpolation did not show any advantages and may therefore be discarded.
The benchmark comparisons with an analytic approach and standard MLPs showed that our framework is able to create trees of MPs with high accuracy, and compared to analytic approaches low inference times.
However, it exhibits slightly higher computation times compared to the benchmark MLPs, due to the increased compuational complexity of evaluating the Gaussian function.
Nonetheless, this concept improves the accuracy, such that there is a trade-off between inference time and precision.

Our approach is able to learn the solutions of arbitrarily complex OCPs such that more complex dynamic models that generate more plausible maneuvers could be employed without affecting the inference time.
It is important to note, that the quality of the trajectories always depends on the OC formulation, and limitations there are also valid for our proposed model.

In our experiments, the basic RBFN model did not converge during training, so the use of the latent space representation was crucial to ensure the convergence.
Finally, our MP-RBFN is differentiable with respect to its parameters by design. This enables future end-to-end training of entire software stacks by integrating MP-RBFN into a fully differentiable PPC framework.

\section{Conclusion and Future Work}
This paper introduces a new generalized formulation of RBFNs for the learning-based generation of vehicle motion primitives. 
The proposed MP-RBFN is a framework to learn basic motion units optimized by arbitrary OCP formulations and thus comply with vehicle dynamics.
The findings highlight the effectiveness of our method in terms of accuracy, computational efficiency, and flexibility. MP-RBFN yields a seven times higher precision than semi-analytic MP calculations and superior inference times compared to OCP solvers.
We showed the applicability in a simulation of an overtaking maneuver of a vehicle with a sampling-based motion planner that utilizes MP-RBFN to generate trajectories.

Future work includes training the MP-RBFN with different robot models and integration into a differentiable PPC stack for end-to-end optimization. The full system will be tested on our EDGAR autonomous vehicle \cite{karle2024edgar} in real-world.

\bibliographystyle{IEEEtran}
\bibliography{refs}

\end{document}